\documentclass[10pt,twocolumn,letterpaper]{article}

\usepackage{cvpr}
\usepackage{times}
\usepackage{epsfig}
\usepackage{graphicx}
\usepackage{amsmath}
\usepackage{amssymb}

\usepackage{array}
\usepackage{eucal}
\usepackage{amsfonts}
\usepackage{color}
\usepackage{epstopdf}
\usepackage{authblk}
\usepackage{enumitem}

\makeatletter
\newcommand{\thickhline}{%
   \noalign {\ifnum 0=`}\fi \hrule height 1pt
   \futurelet \reserved@a \@xhline
}

\newcolumntype{[}{@{\vrule width 1pt\hspace{6pt}}}
\newcolumntype{]}{@{\hspace{6pt}\vrule width 1pt}}
\newcolumntype{!}{@{\hskip\tabcolsep\vrule width 1pt\hskip\tabcolsep}}

\newcommand{\xb}{\mathbf{x}}

\usepackage[breaklinks=true,bookmarks=false,colorlinks,linkcolor=red,pagebackref=true,letterpaper=true]{hyperref}
\usepackage{fancyhdr}
\makeatletter
\newcommand{\printfnsymbol}[1]{%
  \textsuperscript{\@fnsymbol{#1}}%
}
\makeatother

\cvprfinalcopy % *** Uncomment this line for the final submission

\def\cvprPaperID{****} % *** Enter the CVPR Paper ID here

\begin{document}

%%%%%%%%% TITLE
\title{Resolution Adaptive Networks for Efficient Inference}

\newcommand*{\affaddr}[1]{#1} % No op here. Customize it for different styles.
\newcommand*{\affmark}[1][*]{\textsuperscript{#1}}
\author{%
Le Yang\affmark[1]\thanks{Equal Contribution.} \quad Yizeng Han\affmark[1]\printfnsymbol{1}
\quad Xi Chen\affmark[2]\printfnsymbol{1}\thanks{This work is done when Xi Chen was an intern at Tsinghua University.} \quad Shiji Song\affmark[1] \quad Jifeng Dai\affmark[3] \quad Gao Huang\affmark[1]\thanks{Corresponding author}\\
\affaddr{\affmark[1]Tsinghua University, \\Beijing National Research Center for Information Science and Technology (BNRist)\quad}\\
\affaddr{\affmark[2]Harbin Institute of Technology\quad}
\affaddr{\affmark[3]SenseTime}\\
\tt\small{\{yangle15, hanyz18\}@mails.tsinghua.edu.cn, \{shijis, gaohuang\}@tsinghua.edu.cn, xi.chen@stu.hit.edu.cn, daijifeng@sensetime.com}\\
% \affaddr{\LaTeX\ University}%
}
\maketitle
%\thispagestyle{empty}

%%%%%%%%% ABSTRACT
\begin{abstract}
   Adaptive inference is an effective mechanism to achieve a dynamic tradeoff between accuracy and computational cost in deep networks. Existing works mainly exploit architecture redundancy in network depth or width. In this paper, we focus on spatial redundancy of input samples and propose a novel Resolution Adaptive Network (RANet), which is inspired by the intuition that low-resolution representations are sufficient for classifying ``easy'' inputs containing large objects with prototypical features, while only some ``hard'' samples need spatially detailed information. In RANet, the input images are first routed to a lightweight sub-network that efficiently extracts low-resolution representations, and those samples with high prediction confidence will exit early from the network without being further processed. Meanwhile, high-resolution paths in the network maintain the capability to recognize the ``hard'' samples. Therefore, RANet can effectively reduce the spatial redundancy involved in inferring high-resolution inputs. Empirically, we demonstrate the effectiveness of the proposed RANet on the CIFAR-10, CIFAR-100 and ImageNet datasets in both the anytime prediction setting and the budgeted batch classification setting.
   \vskip -0.2in
\end{abstract}

%%%%%%%%% BODY TEXT
\section{Introduction}
   \label{sec-intro}
   Although advances in computer hardware have enabled the training of very deep convolutional neural networks (CNNs), such as ResNet \cite{he2016resnet} and DenseNet \cite{huang2017densenet}, the high computational cost of deep CNNs is still unaffordable in many applications. Many efforts have been made to speed up the inference of deep models, \textit{e.g.}, lightweight network architecture design \cite{howard2017mobilenetsv1,sandler2018mobilenetv2,zhang2018shufflenet,huang2018condensenet}, network pruning \cite{lecun1990optimal,li2016pruning,liu2017learning} and weight quantization \cite{hubara2016bnn,rastegari2016xnor,jacob2018quantization}. Among them, the adaptive inference scheme \cite{lin2017runtime,veit2018AdpGraph,huang2017msdnet,teja2018hydranets}, which aims to reduce the computational redundancy on ``easy'' samples by dynamically adjusting the network structure or parameters conditioned on each input, has been shown to yield promising performance.

   \begin{figure}
   \begin{center}
      \includegraphics[width=2.9in]{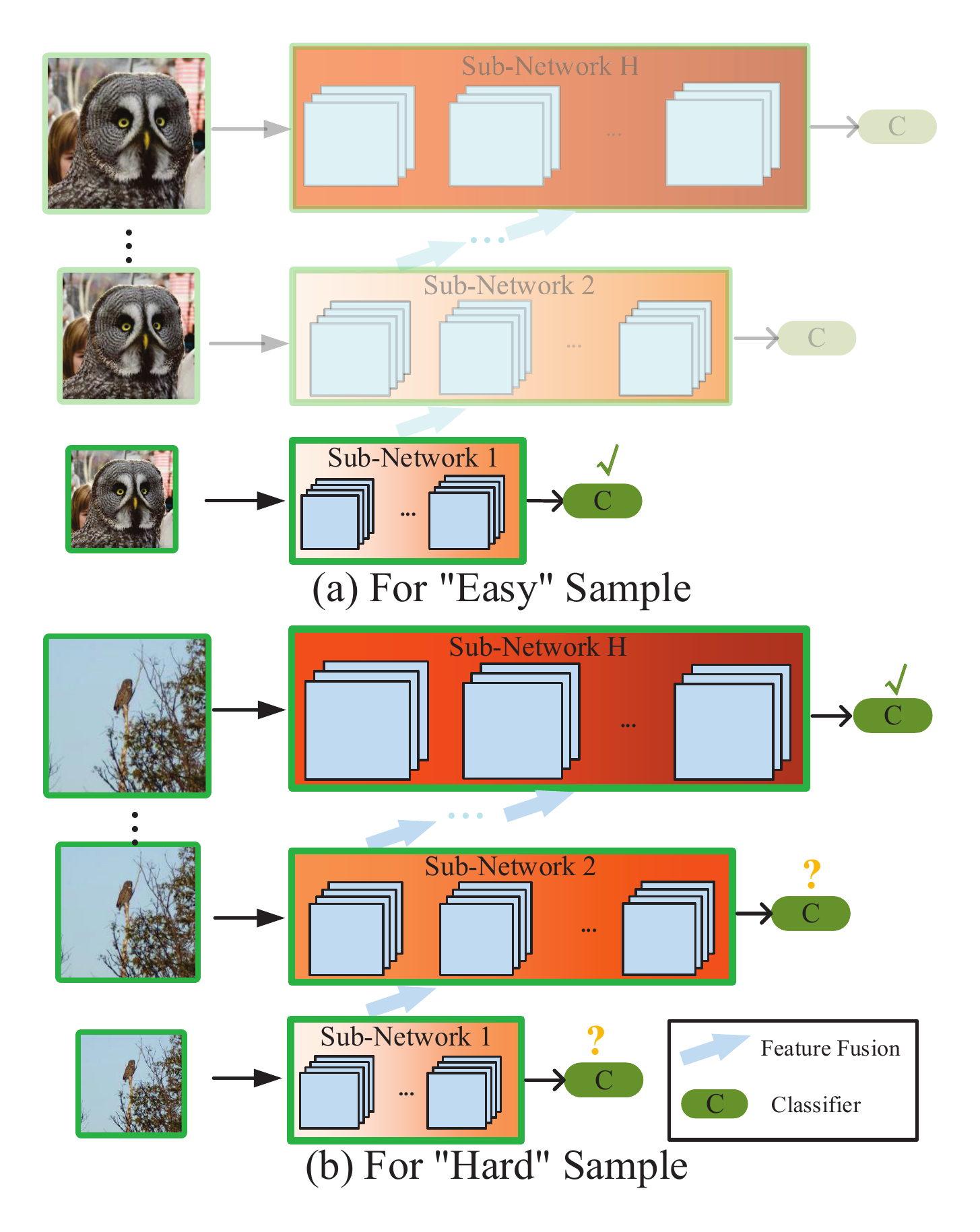}
   \end{center}
   \vskip -0.15in
      \caption{
   Classifying images of owls. In (a), the canonical sample can be recognized by the Sub-network 1 with the lowest resolution, and thus the following sub-networks will be unused. For the ``hard'' image in (b), the Sub-network 1 fails to provide a reliable prediction. Therefore, classifying this sample requires computationally more expensive sub-networks with finer features.
   }
   \vskip -0.15in
   \label{fig1}
   \end{figure}

   Most existing works on adaptive inference focus on reducing the network depth or width for images with easily recognizable features. It has been shown that the intrinsic classification difficulty for different samples varies drastically: some of them can be correctly classified by smaller models with fewer layers or channels, while some may need larger networks \cite{lin2017runtime,veit2018AdpGraph,huang2017msdnet,teja2018hydranets}. By exploiting this fact, many works have been proposed recently. For example, the model in \cite{lin2017runtime} executes runtime pruning of convolutional kernels with a policy learned by reinforcement learning strategies. The network in \cite{veit2018AdpGraph} inserts a linear layer before each convolutional layer to generate a binary decision on whether executing the following convolutional operation dynamically. Multi-Scale Dense Network (MSDNet) \cite{huang2017msdnet} allows some samples to exit at some auxiliary classifiers conditioned on their prediction confidence.

   % and \cite{figurnov2017spatially} adaptively skip layers in a residual block of ResNets\cite{he2016resnet} according to a halting score updated after every convolutional operation.

   In this paper, we consider adaptive inference from a novel perspective. In contrast to existing works focusing on the computational redundancy in the \emph{network structure}, we aim to exploit the information redundancy in the \emph{data samples}. Our motivation is that low-resolution feature representations are sufficient to classify ``easy'' samples (as shown in the top row in \figurename~\ref{fig1}), while applying high-resolution feature maps to probe the details is necessary for accurately recognizing some ``hard'' samples (as shown in the bottom row in \figurename~\ref{fig1}). This further agrees with the ``coarse to fine processing'' efficient algorithm design in \cite{ke2017multigrid}. From a signal frequency viewpoint \cite{chen2019octavenet}, ``easy'' samples could be correctly classified with low-frequency information contained in low-resolution features. High-frequency information is only utilized as complementary for recognizing ``hard'' samples when we fail to precisely predict the samples with low-resolution features.

   Based on the above intuition, we propose a \textit{Resolution Adaptive Network} (RANet) that implements the idea of performing resolution adaptive learning in deep CNNs. \figurename~\ref{fig1} illustrates the basic idea of RANet. It is composed of sub-networks with different input resolutions. The ``easy'' samples are classified by the sub-network with the feature maps in the lowest spatial resolution. The sub-networks with higher resolution will be applied when the previous sub-network fails to achieve a given criterion\footnote{In this paper, we use the prediction confidence from the \emph{softmax} probability.}. Meanwhile, the coarse features from the previous sub-network will be reused and fused into the current sub-network. The adaptation mechanism of RANet reduces computational budget by avoiding performing unnecessary convolutions on high-resolution features when samples can be accurately predicted with low-resolution representations, leading to improved computational efficiency.

   We evaluate the RANet on three image classification datasets (CIFAR-10, CIFAR-100, and ImageNet) under the anytime classification setting and the budgeted batch classification setting, which are introduced in \cite{huang2017msdnet}. The experiments show the effectiveness of the proposed method in adaptive inference tasks.
%-------------------------------------------------------------------------
\section{Related work}
   \textbf{Efficient inference for deep networks.} Many previous works explore variants of deep networks to speed up the network inference. One direct solution is designing lightweight models, \textit{e.g.}, MobileNet \cite{howard2017mobilenetsv1,sandler2018mobilenetv2}, ShuffleNet \cite{zhang2018shufflenet, ma2018shufflenet} and CondenseNet \cite{huang2018condensenet}. Other lines of research focus on pruning redundant network connections \cite{lecun1990optimal,li2016pruning,liu2017learning}, or quantizing network weights \cite{hubara2016bnn,rastegari2016xnor,jacob2018quantization}. Moreover, knowledge distilling \cite{hinton2015distilling} is proposed to train a small (student) network which mimics outputs of a deeper and/or wider (teacher) network.

   The aforementioned approaches can be seen as static model acceleration techniques, which infer all input samples with a whole network consistently. In contrast, adaptive networks can strategically allocate appropriate computational resources for classifying input images based on input complexity. This research direction is gaining increasing attention in recent years due to its advantages. The most intuitive implementation is ensembling multiple models and selectively executing a subset of the models in a cascading \cite{bolukbasi2017adaptive} or mixing way \cite{shazeer2017outrageously, ruiz2019adaptative}. Recent works also propose to adaptively skip layers or blocks \cite{figurnov2017spatially,veit2018AdpGraph,wang2018skipnet,wu2018blockdrop}, or dynamically select channels \cite{lin2017runtime,cai2019model,bejnordi2019batch} during inference time. Auxiliary predictors can also be attached at different locations of a deep network to allow early exiting ``easy'' examples \cite{teerapittayanon2016branchynet,huang2017msdnet, hu2019learning,li2019improved}.
   Furthermore, dynamically activating parts of network branches with multi-branch structure \cite{teja2018hydranets} also provide an alternate way for adaptive inference.

   However, most of these prior works focus on designing adaptive networks by exploiting architecture redundancy of networks. As spatial redundancy of input images has been certificated in recent work \cite{chen2019octavenet}, this paper proposes a novel adaptive learning model which exploits both structural redundancy of a neural network and spatial redundancy of input samples.

   \textbf{Multi-scale feature maps and spatial redundancy. }
    As the downsampling operation in networks with a single scale \cite{he2016resnet,huang2017densenet} may restrict the networks' ability to recognize an object in an arbitrary scale, recent studies propose to adopt multi-scale feature maps in a network to simultaneously utilize both coarse and fine features, which significantly improves the network performance in many vision tasks, including image classification\cite{ke2017multigrid}, object detection \cite{lin2017feature}, semantic segmentation \cite{zhao2018icnet} and pose estimation \cite{sun2019HRNET}. Moreover, the multi-scale structure shows a promising ability in adaptive inference \cite{huang2017msdnet} and memory-efficient network \cite{veniat2018learning}.

   While keeping high-resolution feature maps through a deep neural network is found to be necessary for recognizing some atypical ``hard'' samples or some specific tasks such as pose estimation \cite{sun2019HRNET}, frequently operating convolutions on high-resolution features usually results in resource-hungry models. It has been observed that lightweight networks can yield a decent error rate for all samples with low-resolution inputs \cite{howard2017mobilenetsv1}. The spatial redundancy in these convolutional neural networks has also been studied in \cite{chen2019octavenet}, where the octave convolution in the network processing feature maps with small scales improves the computational efficiency and the classification performance simultaneously. Moreover, ADASCALE proposed in~\cite{chin2019adascale} also adaptively selects the input image scale that improves both accuracy and speed for video object detection.

   \begin{figure*}
      \vskip -0.15in
      \begin{center}
         \includegraphics[width=\textwidth]{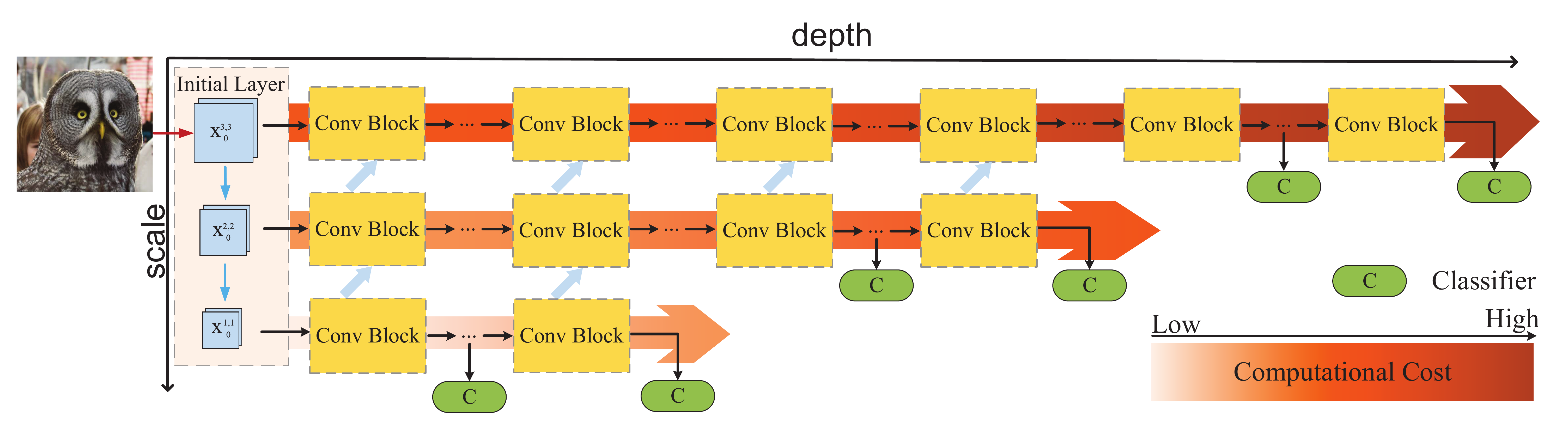}
      \end{center}
         \caption{The illustration of an RANet with three scales. Classifiers only operate on feature maps at the lowest resolution.}
      \label{fig_arch}
      \end{figure*}
   However, none of these existing works considers designing an adaptive model by exploiting spatial redundancy in images. In this paper, we propose our RANet for resource-efficient image classification, motivated by the intuition that a smaller scale can be capable of handling most of input samples. Compared to ADASCALE~\cite{chin2019adascale}, which also adaptively selects the input image scale for vision task, the proposed RANet can be implemented for the budgeted classification setting during adaptive inference. Our work achieves resolution adaptation by classifying some of inputs on small scales and allowing larger scales to be processed only when inputs can not be recognized with coarse representations. The resolution adaptation in RANet significantly improves its computational efficiency without sacrificing accuracy.

\section{Method}
In this section, we first introduce the idea of adaptive inference, then we demonstrate the overall architecture and the network details of our proposed RANet.

\subsection{Adaptive Inference Setting}
We set up an adaptive inference model as a network with $K$ classifiers, where these intermediate classifiers are attached at varying depths of the model. Given an input image $\xb$, the output of the $k$-th classifier ($k \!=\! 1,\cdots,K$) can be represented by
\begin{align}
   \mathbf{p}^k = f_k(\xb;\theta_k) = [p^k_1,\cdots,p^k_C]^\mathrm{T} \in \mathbb{R}^C,
\end{align}
where $\theta_k$ denotes the parameters of the partial network corresponding to the $k$-th classifier, and each element $p^k_c\in\left[0,1\right]$ is the prediction confidence for the $c$-th class. Note that $\theta_k$'s have shared parameters here.

The adaptive model infers a sample by dynamically allocating appropriate computational resources depending on the complexity of this sample. A sample will exit the network at the first classifier whose output satisfies a certain criterion. In this paper, we use the highest confidence of the \emph{softmax} output as our decision basis, which means that the final output will be the prediction of the first classifier whose largest \emph{softmax} output is greater than a given threshold $\epsilon$. This can be represented by
{\setlength\abovedisplayskip{4pt}
 \setlength\belowdisplayskip{4pt}
\begin{align}
&k^* = \min \left\{k |  \max_c\ \  p^k_c \geq \epsilon \right\}, \\
&\hat{y}\ \in arg\max_c  \ \ \  p^{k^*}_c.
\end{align}
}The threshold $\epsilon$ controls the trade-off between classification accuracy and computational cost at test time.

\subsection{Overall Architecture}
\figurename~\ref{fig_arch} illustrates the overall architecture of the proposed RANet. It contains an \textit{Initial Layer} and $H$ sub-networks corresponding to different resolutions. Each sub-network has multiple classifiers at the last few blocks. Similar to MSDNet \cite{huang2017msdnet}, we adopt a multi-scale architecture and dense connection in our approach. Although RANet and MSDNet have a similar multi-scale structure, their detailed architecture designs and computation graphs differ significantly. The most prominent difference is that RANet needs to extract \emph{low-resolution features first}, which does not follow the traditional design routine in classical deep CNNs (including MSDNet, ResNet, DenseNet, etc.) that all extract \emph{high-resolution features first}. More details of the differences between MSDNet and our RANet will be discussed in Section \ref{differences}.

The basic idea of RANet is that the network will first predict a sample with the first sub-network, using feature maps of the lowest spatial resolution to avoid the high computational cost induced by performing convolutions on large scale features. If the first sub-network makes an unreliable prediction of the sample, the small scale intermediate features will be fused into the next sub-network with a higher resolution. The classification task is then conducted by the next sub-network with larger scale features. This procedure is repeated until one sub-network yields a confident prediction, or the last sub-network is utilized.

The adaptive inference procedure of RANet is further illustrated in \figurename~\ref{fig_arch}: with $H$ sub-networks ($H\!\!=\!\!3$ in the illustration) and an input sample $\xb$, the network will first generate $H$ base feature maps in $S$ scales (For instance, there are 3 scales in the illustration, and $s\!\! = \!\!1$ represents the lowest resolution). The base features in scale $s$ corresponding to Sub-network $h$ can be denoted as $\xb^{s,h}_0, s\!\!= \!\!1,2,...S, h\!\!= \!\!1,2,...H$. Then the classification task is first conducted by Sub-network 1 using features $\xb^{1,1}_0$ at the bottom. If Sub-network 1 fails to achieve the classification result with a high confidence, Sub-network 2, which processes larger scale features ($\xb^{2,2}_0$), will be utilized for further classifying the sample. The intermediate features in Sub-network 1 are successively fused into Sub-network 2. We repeat this procedure for Sub-network 3 if Sub-network 2 fails to make a confident prediction.

It is worth noting that even RANet processes inputs from coarse to fine in general, each sub-network in RANet still downsamples features during forward propagation until reaching the lowest resolution ($s\!\!=\!\!1$), and all the classifiers are only attached at the last few blocks with $s\!\!=\!\!1$ in each sub-network.

The aforementioned inference procedure meets our intuition for image recognition. An ``easy'' sample with representative characteristics can be correctly classified sometimes with high confidence even only low-resolution representations are provided. A ``hard'' sample with atypical features can only be correctly recognized based on global information accompanied with fine details, which are extracted from high-resolution feature maps.

\subsection{Network Details}
This subsection provides more detailed introductions about each component in RANet.
\vspace{-0.5em}
\subsubsection{Initial Layer}
An \textit{Initial layer} is implemented to generate $H$ base features in $S$ scales and it only includes vertical connections in \figurename~\ref{fig_arch}.  One could view its vertical layout as a miniature ``$H$-layers'' convolutional network ($H$ is the number of base features in the network). \figurename~\ref{fig_arch} shows an RANet with 3 base features in 3 scales. The first base features with the largest scale is derived from a \textit{Regular-Conv} layer\footnote{A \textit{Regular-Conv} layer in this paper is consisted of a bottleneck layer and a regular convolution layer. Each layer is composed of a Batch normalization (BN) layer \cite{ioffe2015BatchNorm}, a ReLU layer \cite{nair2010relu} and a convolution layer.}, and the coarse features are obtained via a \textit{Strided-Conv} layer\footnote{A \textit{Strided-Conv} layer is realized by setting the stride of the second convolution in \textit{Regular-Conv} layer as 2.} from the former higher-resolution features. It is worth noting that the scales of these base features can be the same. For instance, one could have an RANet with 4 base features in 3 scales, where the scales of the last two base features are of the same resolution.
\vspace{-0.5em}
\subsubsection{Sub-networks with Different Scales}
As the \textit{Initial layer} generates $H$ base features, the proposed network can then be separated into $H$ sub-networks, which are further composed by different \textit{Conv Blocks}. Each sub-network, except the first one, conducts the classification task with its corresponding base feature maps and features from the previous sub-network.

\begin{figure}
\vskip -0.15in
\begin{center}
   \includegraphics[width=3in]{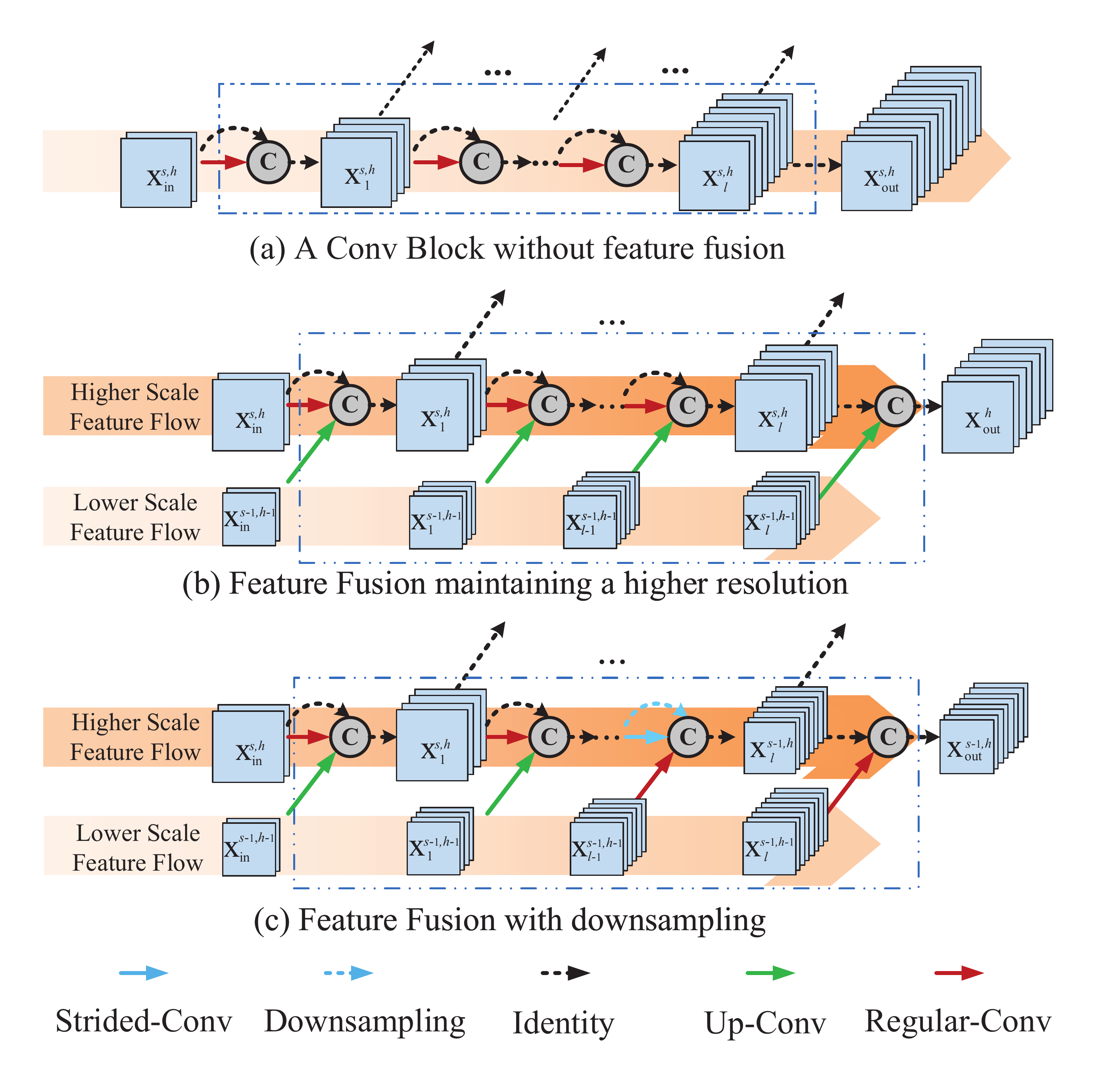}
\end{center}
\vskip -0.1in
   \caption{Two kinds of \textit{Conv Blocks} in RANet: \textit{Dense Block}, (a), and \textit{Fusion Block}, (b,c). Moreover, the block in (b) maintains the input resolution of the feature maps, while the block in (c) downsamples the features by a factor of 2 at the end of the block.}
\label{fig_norm}
\vskip -0.15in
\end{figure}
\textbf{Sub-network 1.}
Sub-network 1 with input $\xb^{1,1}_0$ processes the lowest-resolution features. We adopt regular \textit{Dense Blocks} \cite{huang2017densenet} with $l$ layers in Sub-network 1, which is shown in \figurename~\ref{fig_norm} (a). Moreover, the $i$-th layer's output $\xb^{1,1}_i,i\!\!= \!\!1,2,...l$ in each \textit{Dense Block} is also propagated to Sub-network 2 to reuse the early features. In general, one can view Sub-network 1 as a DenseNet with multiple classifiers, processing the lowest-resolution feature maps.

\textbf{Sub-networks on larger-scale features.}
Sub-network $h$ ($h\!>\!1$) with scale $s$ processes the base features $\xb^{s,h}$ and fuses the features from Sub-network ($h\!-\!1$). We call \textit{Conv Blocks} with feature fusion as \textit{Fusion Blocks} (shown in \figurename~\ref{fig_norm} (b, c)). Suppose that Sub-network ($h\!-\!1$) has $b_{h-1}$ blocks, then the first $b_{h-1}$ blocks in Sub-network $h$ will all be \textit{Fusion Blocks}.

We design two different ways of feature fusion. One maintains the input resolution, which is illustrated in \figurename~\ref{fig_norm} (b), while the other reduces the feature scale by a \textit{Strided-Conv} layer, as shown in \figurename~\ref{fig_norm} (c). To generate new feature maps with higher resolution as inputs, the \textit{Fusion Block} in \figurename~\ref{fig_norm} (b) first produces $\xb^{s,h}_{\text{in}}$ with a \textit{Regular-Conv} layer. Features in scale $(s\!-\!1)$ from the previous sub-network is processed by an \textit{Up-Conv} layer, which is composed of a \textit{Regular-Conv} layer and an up-sampling bilinear interpolation. This ensures the produced features are of the same spatial resolution. The resulting features are then fused through concatenation with dense connection.

As shown in \figurename~\ref{fig_norm} (c), a \textit{Fusion Block} with downsampling utilizes a \textit{Strided-Conv} layer to reduce the spatial resolution at the end of the block. Concatenation with dense connection is also conducted after a pooling operation as shown by a blue dashed arrow. Since the feature scale is reduced in the current sub-network, features from the previous sub-network are processed by a \textit{Regular-Conv} layer to maintain the low resolution, and then fused by concatenation at the end of the block in \figurename~\ref{fig_norm} (c).

Sub-network $h$ with scale $s$  can be established as follow: for a sub-network with $b_h$ blocks, block 1 to block $b_{h-1}$ ($b_{h-1}\!\!<\!\!b_n$) are all \textit{Fusion Blocks}, while the rest of them are regular \textit{Dense Blocks}. Moreover, we downsample the feature maps $s$ times at the $b_{h-s}$,...,$b_{h-1}$-th blocks during forward propagation. This ensures that at the end of each sub-network where we attach classifiers, the features must be of the lowest resolution.

\textbf{Transition layer.}  Similar to the architecture design in \cite{huang2017densenet} and \cite{huang2017msdnet}, we implement \textit{Transition layers} to further compress the feature maps in each sub-network. The design of a \textit{Transition layer} is exactly the same as the one in \cite{huang2017densenet} and \cite{huang2017msdnet}, which is composed of a $1\times1$ convolution operator following by a BN layer and a ReLU layer. \textit{Transition layers} further guarantee the computational efficiency of the proposed network. For simplicity, we omit these \textit{Transition Layers} in \figurename~\ref{fig_arch}.

\textbf{Classifiers and loss function.} The classifiers are implemented at the last few blocks of different sub-networks. At the training stage, we let input samples pass through Sub-network 1 to Sub-network $H$ sequentially and cross-entropy loss function is used for each classifier. We set the overall loss function for RANet as a weighted cumulative loss of these classifiers. We empirically follow the settings in \cite{huang2017msdnet} and use the same weight for all loss functions in this paper.

\subsection{Resolution and Depth Adaptation}\label{differences}
\begin{figure}
   \begin{center}
      \includegraphics[width=0.45\textwidth]{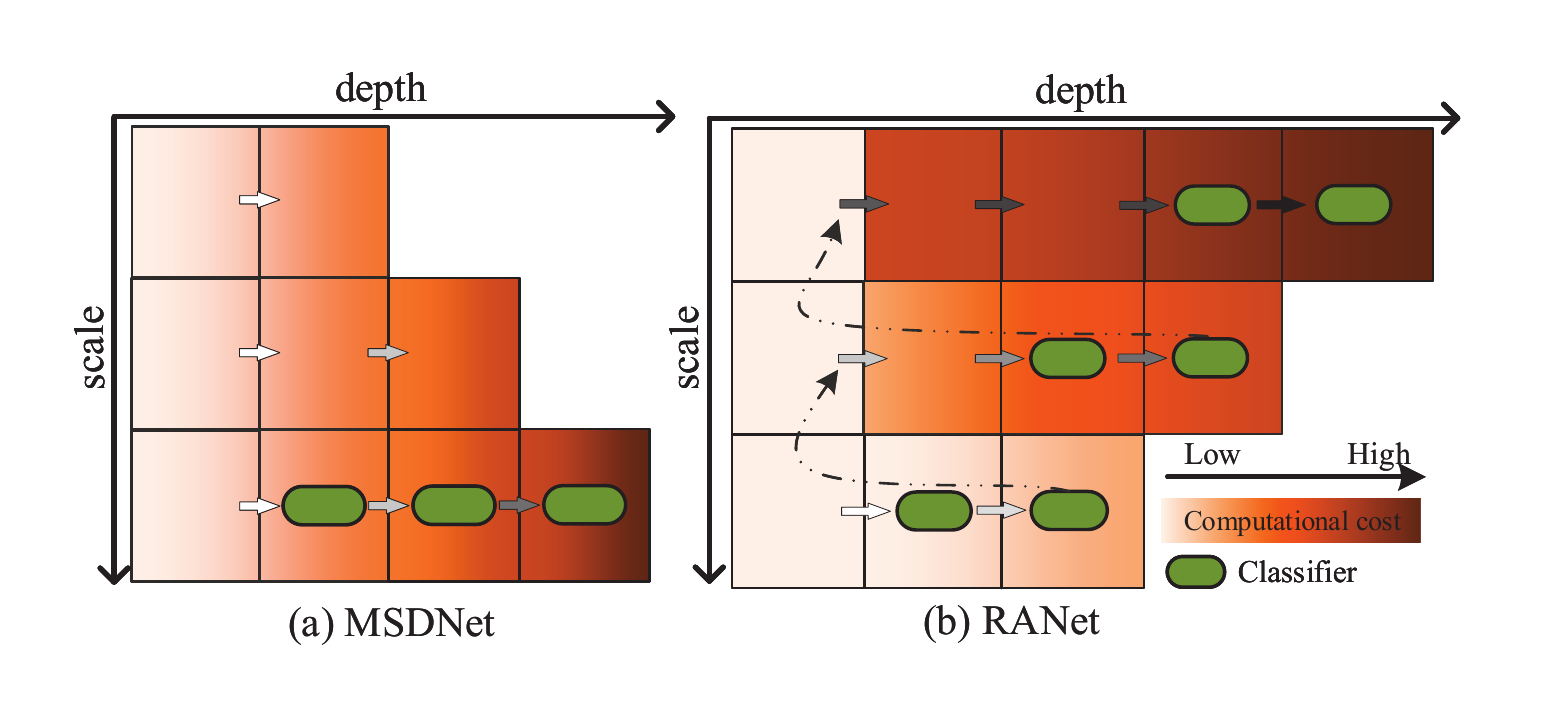}
   \end{center}
   \vskip -0.1in
      \caption{Depth adaptation in MSDNet (a) and resolution-depth adaptation in our RANet (b). Different shaded areas represent the network blocks with varied computational costs, and the colored arrows represent the feature propagation path. The lighter the color is, the earlier the propagation is executed.  The dashed arrows in (b) indicate that RANet adopts a zigzag-shape computation graph from the bottom to the top.}
   \label{fig_compare}
\vskip -0.2in
   \end{figure}

Our proposed RANet can simultaneously implement the idea of depth adaptation, which is adopted in MSDNet \cite{huang2017msdnet}, and resolution adaptation. \figurename~\ref{fig_compare} illustrates the main differences between MSDNet (left) and our RANet (right). In MSDNet, the classifiers are located at the lowest resolution scale, and once an intermediate predictor does not yield a confident prediction, the following layers of all scales will be executed. However, in our RANet, the \textit{Dense Blocks} with the smallest scale input are first activated sequentially and the depth adaptation is conducted within a single scale. If the previous sub-network cannot make a confident prediction, the input sample will be propagated to the next sub-network and repeat the depth adaptation process until the prediction confidence meets the criterion, or the last classifier of the whole network is reached. Such an inference scheme naturally combines resolution and depth adaptation, achieving significant improvement over MSDNet.

\section{Experiments}
To demonstrate the effectiveness of our approach, we conducted experiments on the CIFAR \cite{krizhevsky2009cifar} and ImageNet \cite{deng2009imagenet} datasets. The code is available at \url{https://github.com/yangle15/RANet-pytorch}. The implementation details of RANets and MSDNets in our experiments are described in Appendix A.

\textbf{Datasets. }The CIFAR-10 and CIFAR-100 datasets contain $32\times32$ RGB natural images, corresponding to 10 and 100 classes, respectively. The two datasets both contain 50,000 training and 10,000 testing images. Following \cite{huang2017msdnet}, we hold out 5,000 images in the training set as a validation set to search the optimal confidence threshold for adaptive inference. The ImageNet dataset contains 1.2 million images of 1,000 classes for training, and 50,000 images for validation. For adaptive inference tasks, we use the original validation set for testing, and hold out 50, 000 images from the training set as a validation set.

\textbf{Training policy. }We train the proposed models using stochastic gradient descent (SGD) with a multi-step learning rate policy. The batch size is set to 64 and 256 for the CIFAR and ImageNet datasets, respectively. We use a momentum of 0.9 and a weight decay of $1\times10^{-4}$. Moreover, for the CIFAR datasets, the models are trained from scratch for 300 epochs with an initial learning rate of 0.1, which is divided by a factor of 10 after 150 and 225 epochs. The same training scheme is applied to the ImageNet dataset. And we train the models for 90 epochs from scratch and the initial learning rate decreases after 30 and 60 epochs.

\textbf{Data augmentation. } We follow \cite{he2016resnet} and apply standard data augmentation schemes on the CIFAR and ImageNet datasets. On the two CIFAR datasets, images are randomly cropped to samples with $32\times32$ pixels after zero-padding (4 pixels on each side). Furthermore, images are horizontally flipped with probability 0.5 and RGB channels are normalized by subtracting the corresponding channel mean and divided by their standard deviation. On ImageNet, we follow the data augmentation scheme in \cite{he2016resnet} for training, and apply a $224\times224$ center crop to images at test time.

\subsection{Anytime Prediction}
In the anytime prediction setting \cite{huang2017msdnet}, we evaluate all classifiers in an adaptive networks and report their classification accuracies with corresponding FLOPs (floating point operations).

\textbf{Baseline models. } Following the setting in \cite{huang2017msdnet}, in addition to MSDNet, we also evaluate several competitive models as our baselines, including $\text{ResNet}^\text{MC}$, $\text{DenseNet}^\text{MC}$ \cite{lee2015MC}, and ensembles of ResNets and DenseNets of \textit{varying} sizes. Details on architectural configurations of MSDNets and RANets in the experiments are described in Appendix A. As recent research in \cite{li2019improved} investigates improved techniques for training adaptive networks, we further evaluate these techniques on both RANet and MSDNet. The experiments show that the computational efficiency of the RANet can be further improved and outperforms the improved MSDNet. The results are provide in Appendix B.

% Besides model configurations, training techniques designed for adaptive learning with early exits \cite{li2019improved} can further improve the model performance. We utilize the techniques described in \cite{li2019improved} to enhance our model and the results are provided in Appendix A.

\begin{figure*}
\begin{center}
   \includegraphics[width=\textwidth]{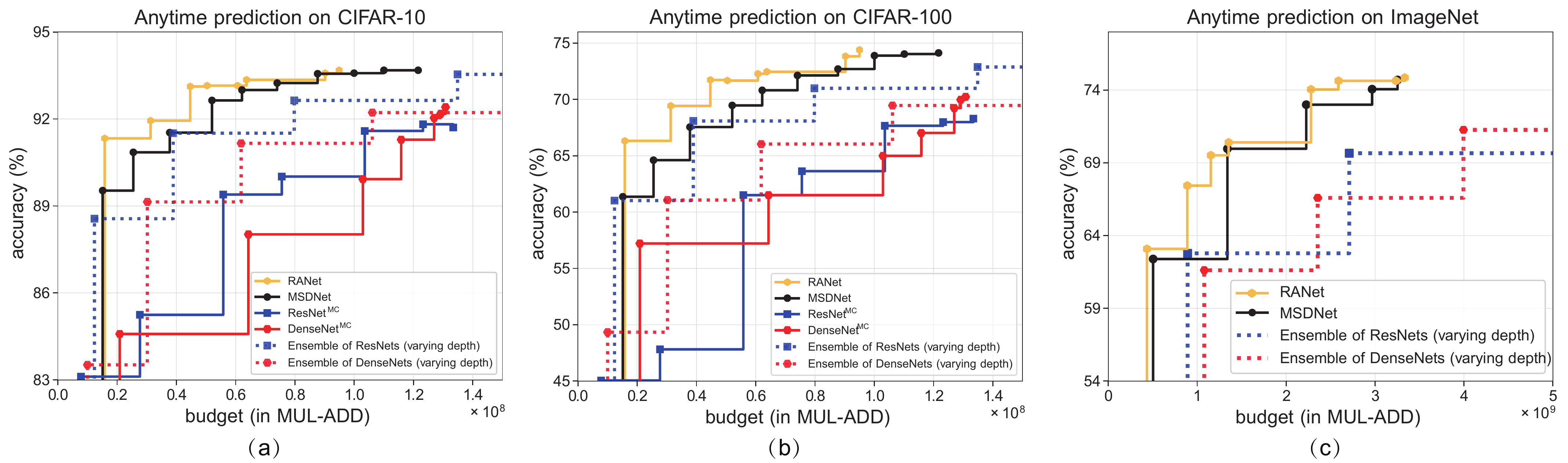}
\end{center}
\vskip -0.1in
   \caption{Accuracy (top-1) of anytime prediction models as a function of computational budget on CIFAR-10 (left), CIFAR-100 (middle) and ImageNet (right). Higher is better.}
   \vskip -0.1in
\label{anytime}
\end{figure*}

\textbf{Results. } We report classification accuracies of all individual classifiers in our model and other baselines. The results are summarized in \figurename~\ref{anytime}. The evaluated MSDNets and RANets are depicted by black and yellow lines, respectively. In general, MSDNet substantially outperforms other baseline models, and RANet are superior to MSDNet, especially when the computational budget is low.

In particular, on CIFAR-10 (CIFAR-100), the accuracies of different classifiers for RANet are over $1\%$ ($2\%\!-\!5\%$) higher than those of MSDNet when the computational budget ranges from $0.1\times10^8$ to $0.5\times10^8$ FLOPs. Moreover, compared to MSDNet, RANet achieves its highest accuracy with less computational demands (around $0.25\times10^8$ FLOPS). On ImageNet, the proposed network outperforms MSDNet by around $1\%\!-\!7\%$ when the budget ranges of $0.5\times10^{9}$ to $1.5\times10^{9}$ FLOPs. Although both MSDNet and RANet achieve similar classification accuracy ($74\%$) at the last classifier, our model only uses around $27\%$ fewer FLOPs compared to MSDNet.

At the first classifier, the accuracies of RANets are $2\%$ and $5\%$  higher than those of MSDNets on CIFAR-10 and CIFAR-100, respectively. On the ImageNet dataset, RANet still slightly outperforms MSDNet at the first classifier. With $1.0\times10^{9}$ FLOPs, RANet can achieve a classification accuracy of around $68\%$, which is around $5\%$ higher than that achieved by MSDNet. We also observe that ensembles of ResNets outperform MSDNets in low-budget regimes, because the predictions of ensembles are performed by the first lightweight networks, which are optimized exclusively for the low budget. However, RANets are consistently superior to ensembles of ResNets on all datasets. This meets our expectation that Sub-network 1 with the first classifier in RANet is specially optimized for recognizing  ``easy'' samples. Since Sub-network 1 directly operates on the feature maps with the lowest resolution, it avoids performing the convolutions on high-resolution feature maps, which results in the high computational efficiency of the first classifier. Furthermore, as Sub-network 1 in RANets can be viewed as exclusively optimized lightweight models, the early classifiers of RANets show their advantages in the classification tasks. Different from ResNet ensembles, which repeat the computation of similar low-level representations, RANets fuse the feature maps from previous lightweight networks into a large network to make full use of the obtained features. This mechanism effectively improves classification accuracies when we have more computational resources.

\subsection{Budgeted Batch Classification}
The budgeted batch classification setting is described in \cite{huang2017msdnet}. We set a series of thresholds that depend on different computational budgets.  For a given input image, we let it pass through each classifier in an adaptive network, sequentially. The forward propagation stops at the classifier whose output confidence reaches the given threshold, and then we report its prediction as the final result for this image.

\textbf{Baseline models. }For CIFAR-10 and CIFAR-100, we use ResNet, DenseNet and DenseNet* \cite{huang2017msdnet} as baseline models. For ImageNet, we additionally evaluate ResNet and DenseNet with multi-classifier \cite{lee2015MC}. Performance of some classical deep models are also reported in the experimental results, such as WideResNet \cite{zagoruyko2016wide} (for CIFAR) and GoogLeNet \cite{szegedy2015googleNet} (for ImageNet). See Appendix A for details about the architecture configurations of MSDNets and RANets in the experiments. Moreover, we implement the techniques in \cite{li2019improved} to further evaluate the improved RANets and MSDNets. The results are provided in Appendix B.

\begin{figure*}
\begin{center}
   \includegraphics[width=\textwidth]{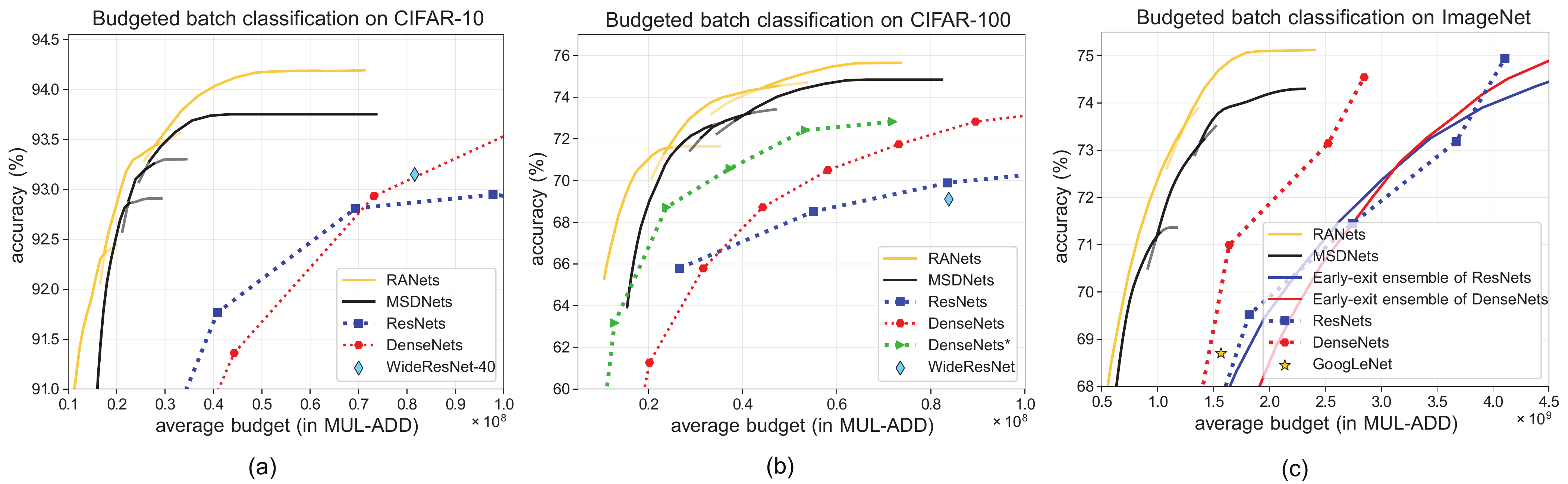}
\end{center}
\vskip -0.1in
   \caption{Accuracy (top-1) of budgeted batch classification models as a function of average computational budget per image on CIFAR-10 (left), CIFAR-100 (middle) and ImageNet (right). Higher is better.}
\vskip -0.05in
\label{dynamic}

\end{figure*}

\textbf{Results. }The results are summarized in \figurename~\ref{dynamic}. We plot the classification accuracy of each MSDNet and RANet in a gray and a light-yellow curve, respectively. We select the best model for each budget based on its accuracy on the test set, and plot the corresponding accuracy as a black curve (for MSDNet) or a golden curve (for RANet).

The results on the two CIFAR datasets show that RANets consistently outperform MSDNets and other baseline models across all budgets. In general, the networks with multi-scale dense connection architecture are always substantially more accurate than other baseline models with the same amount of computation cost under the budgeted batch classification setting. For low computational budget (less than $0.2\times10^8$ FLOPs), on CIFAR-10, the proposed model uses $20\%$ fewer FLOPs to achieve the classification accuracy of $92\%$ compared to MSDNet. On CIFAR-100, RANet can achieve the classification accuracy of $68\%$ with only about $60\%$ FLOPs compared to MSDNet. Even though our model and MSDNet show close performance on CIFAR-10 when the computational budget ranges from $0.2\times10^8$ to $0.3\times10^8$, the classification accuracies of RANets are consistently higher than ($~1\%$) these of MSDNets on CIFAR-100 in median and high budget intervals (over $0.2\times10^8$ FLOPs). Moreover, our model can achieve an accuracy of $94.2\%$ when the budget is higher than $0.2\times10^8$ FLOPs. This accuracy is $~0.5\%$ higher than that of MSDNet under the same computational budget condition. The experiments also show that RANets are up to 4 times more efficient than WideResNets on CIFAR-10 and CIFAR-100.

The experiments on ImageNet yield similar results to those on CIFAR. We observe that RANets consistently surpass MSDNets. Our networks win about $0.5\%$, $1\%$ and $1.2\%$ in terms of top-1 accuracy with $0.75\times10^9$, $1\times10^9$ and $1.75\times10^9$ FLOPs respectively. The results indicate that our RANet outperforms MSDNet by a larger margin as more computational resources are provided. With the same FLOPs, our models achieve more accurate classification results than these popular deep neural networks. With the same classification accuracy, our model reduces the computational budget by around $65\%$, $56\%$ and $44\%$ compared to GoogLeNet, ResNets and DenseNets, respectively. All these results demonstrate that the resolution adaptation along with the depth adaptation can significantly improve the performance of adaptive networks under the budgeted batch classification setting.

\subsection{Visualization and Discussion}
\begin{figure}
\begin{center}
   \includegraphics[width=3.3in]{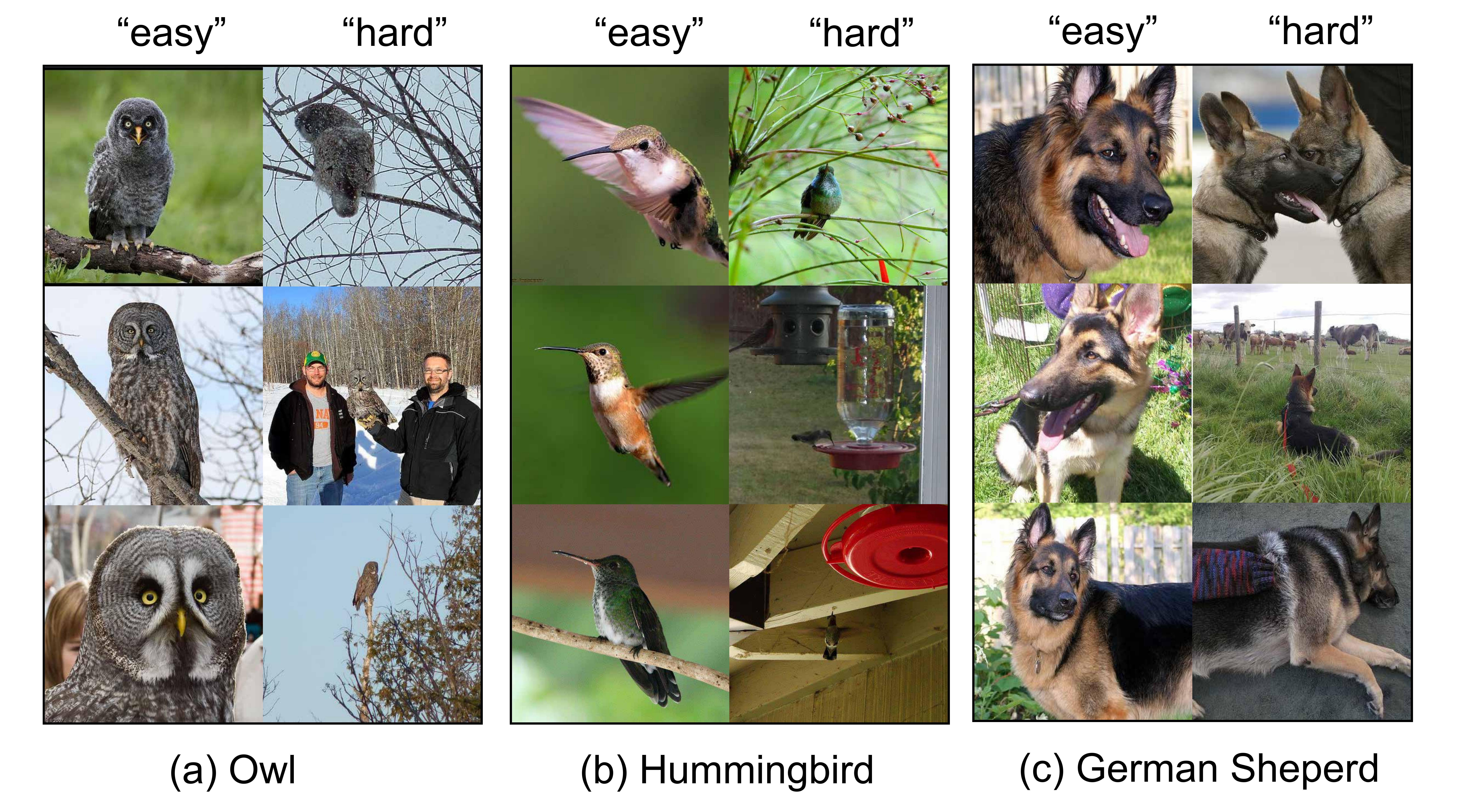}
\end{center}

   \caption{Visualization of ImageNet samples: Owl, Hummingbird and German Shepherd. The column on the left of each sub-figure: the images that exit from the earlier classifiers (``easy'' samples); The column on the right of each sub-figure: the images that fail to be correctly classified at the earlier classifiers but are successfully recognized at the last few classifiers (``hard'' samples).}
   \vskip -0.1in
\label{vis}

\end{figure}

\figurename~\ref{vis} illustrates the ability of RANet to recognize samples with different difficulties. In each sub-figure, the left column shows ``easy'' samples that are correctly classified by the earlier classifiers with high classification confidence.
The right column shows ``hard'' samples that fail to reach sufficient confidence at the early exits and are passed on to the deeper sub-networks handling high-resolution features.
The figure suggests that the earlier classifiers can recognize prototypical samples of a category, whereas the later classifiers are able to recognize non-typical samples, which is similar to the experimental results in \cite{huang2017msdnet}.

It is also observed that the high-resolution feature maps and their corresponding sub-networks are necessary for accurately classifying the object in three different cases.

 $\bullet$\textbf{ Multiple objects.} We find that an image containing multiple objects can be viewed as a ``hard'' sample for RANet. The co-occurrence of different objects may corrupt the feature maps and therefore confuse the early classifiers. In this case, the relationship between each object is a key factor that can seriously affect the categorical prediction of the whole image. For example, in \figurename~\ref{vis} (a), the ImageNet dataset refers the image with an owl on a man's hand as the class ``owl'', even though there are two people in this image. Apparently, rapid downsampling could submerge the own in the image and the network can recognize it as the class ``person''. Furthermore, categorizing this image as the class ``owl'' may result from human perception that we consider objects on a person's hand at the center of an image as more important information. This complex relationship can only be exploited with stronger representations learned by a powerful network.

 $\bullet$\textbf{ Tiny objects.} It is observed that the images with tiny objects always pass through the whole network and thus are also considered as ``hard'' samples for RANet. A possible explanation for this phenomenon is that the information of these tiny target objects in the images can be completely lost after rapidly downsampling the images. The clues for classifying those tiny objects can only be obtained by processing the high-resolution feature maps. For instance, in the right image on the second row of \figurename~\ref{vis} (b), the hummingbird drinking water is too small. Therefore, the representations of the hummingbird can easily be lost due to the rapid downsample operations and might be completely vanished in the coarse feature maps. This makes the image unable to be recognized until the high-resolution feature maps are used for inference, which results in its late exiting in our adaptive inference network.

 $\bullet$\textbf{ Objects without representative characteristics.} Another kind of ``hard'' samples for RANet contain objects without representative characteristics. Such samples are not uncommon due to various factors (such as lighting conditions and shooting angles). In this scenario, we conjecture that the network learns to utilize alternative characteristics instead of representative ones for image recognition. For instance, by comparing the ``easy'' and ``hard'' samples in \figurename~\ref{vis} (c), the network can easily recognize the German Shepherd as long as its facial features are presented completely in the images. However, without complete facial features, a German Shepherd can only be correctly classified at the last classifier. For those ``hard'' samples, the network may take the fur texture of the German Shepherd as the alternative discriminative features during inference. Therefore, without complete facial information, the network learns to correctly classify German Shepherd by searching useful alternative characteristics in high-resolution feature maps.

The rationality and effectiveness of the resolution adaptation can be further understood from the signal frequency perspective, which has been demonstrated and verified in \cite{chen2019octavenet}.  The low-frequency information encoded in low-resolution features, which usually contains global information, can be sufficient for successful classification of most input samples. Nevertheless, higher frequencies encoded with fine details are obligatory for classifying those untypical samples.

%------------------------------------------------------------------------
\vspace{-0.1em}
\section{Conclusion}
\vspace{-0.1em}
In this paper, we proposed a novel resolution adaptive neural network based on a multi-scale dense connection architecture, which we refer to as \textit{RANet}.  RANet is designed in a way that lightweight sub-networks processing coarse features are first utilized for image classification. Samples with high prediction confidence will exit early from the network and larger scale features with finer details will only be further utilized for those non-typical images which achieve unreliable predictions in previous sub-networks. This resolution adaptation mechanism and the depth adaptation in each sub-network of RANet guarantee its high computational efficiency. On three image classification benchmarks, the experiments demonstrate the effectiveness of the proposed RANet in both the anytime prediction setting and the budgeted batch classification setting.

\vspace{-0.07in}
\section*{Acknowledgment}
\vspace{-0.06in}
This work is supported by grants from the Institute for Guo Qiang of Tsinghua University,  National Natural Science Foundation of China (No. 61906106) and Beijing Academy of Artificial Intelligence (BAAI).

{
   \small
   \bibliographystyle{ieee_fullname}
   \bibliography{RANetbib_abb}
}

\end{document}

\title{Supplementary Materials for: Resolution Adaptive Networks for Efficient Inference}
\author{}
\date{}
\maketitle

\section{Appendix A: Implementation Details}
\begin{figure*}
    \begin{center}
    \includegraphics[width=\textwidth]{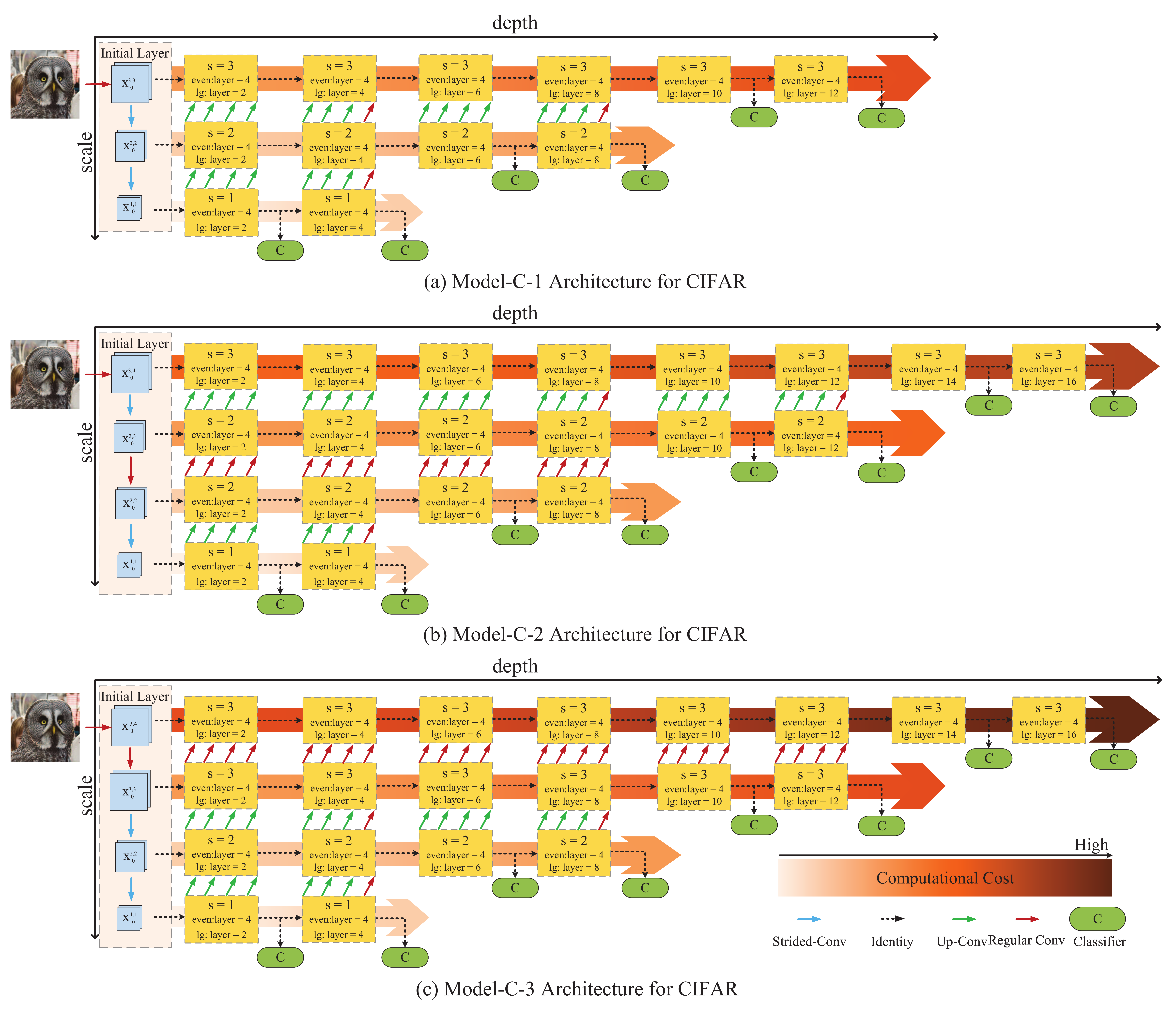}
    \end{center}
    \vskip -0.2in
    \caption{
        Architecture of RANets for CIFAR-10 and CIFAR-100.
    }
    \label{fig_cifar}
\end{figure*}

\begin{figure*}
    \begin{center}
    \includegraphics[width=\textwidth]{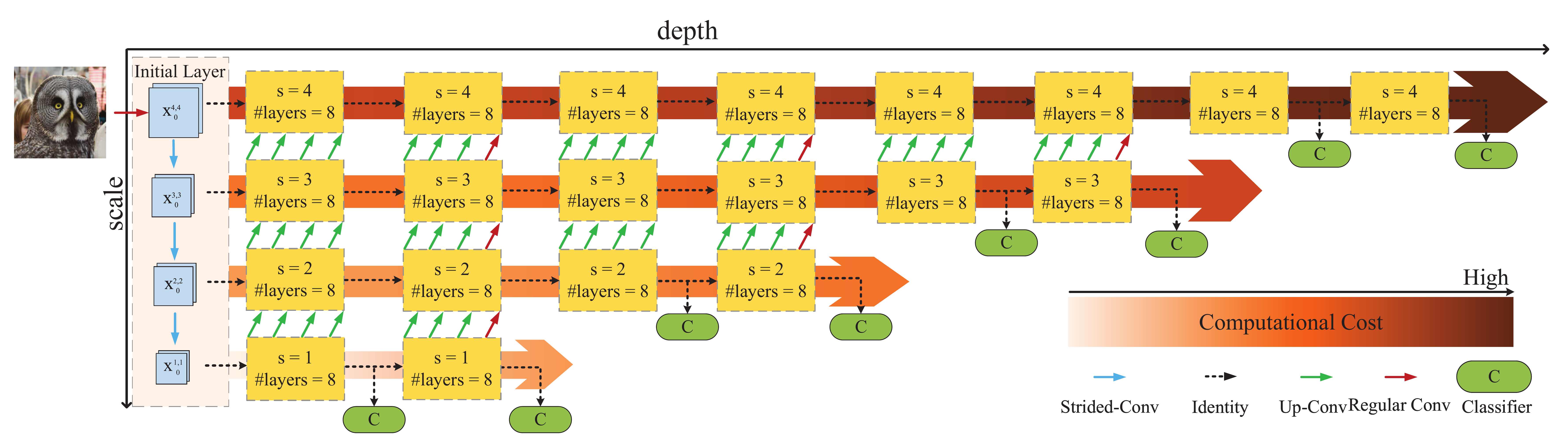}
    \end{center}
    \vskip -0.1in
    \caption{
       Architecture of RANets for ImageNet.
    }
    \label{fig_ImageNet}
\end{figure*}
In this section, we introduce the architecture configurations for our RANets and MSDNets in the experiments of the main paper.
\subsection{CIFAR-10 and CIFAR-100}
\paragraph{MSDNet: } For CIFAR-10 and CIFAR-100, features with 3 different  scales ($32 \times 32$, $16\times 16$, $8\times 8$) are used for MSDNets in our experiments. The trained MSDNets have $\{6, 8, 10\}$ classifiers, where their depths are $\{16, 20, 24\}$, respectively.

\paragraph{RANet: } The same 3 scales features are utilized for our RANets in the experiments. However, as mentioned in section 3.3.1, different from MSDNet, the scales of the generated base features can be different, and we could have a RANet with three or four base features in three scales. We test 3 architecture configurations as follows:

\textbf{Model-C-1}: The size of three base features are $32 \times 32, 16\times 16, 8\times 8$. Three sub-networks corresponding to these base features have $6,4,2$ \textit{Conv Blocks}, respectively. We set two step mode for RANet to control the number of layers in each \textit{Conv Block}: 1)even: the number of layers in each \textit{Conv Block} is set to $4$; 2)linear growth (lg): the number of layers in a \textit{Conv Block} is added $2$ to the previous one, and the base number of layers is $2$. The channel numbers in these base features are $16, 32, 64$, which are input channels numbers for different sub-networks. The growth rates of the 3 sub-networks are $6, 12, 24$. Moreover, for each \textit{Fusion Block}, a compress factor of $0.25$ is applied, which means that $75\%$ of the new added channels are generated from the current sub-network and the other $25\%$ are calculated from the previous sub-network with lower feature resolution. Furthermore, we add $s$ transition layers for Sub-network $s$. E.g., we add one 3 transition layers for Sub-network 3. The Model-C-1 has six classifiers in total, and its overall architecture is illustrated in Figure \ref{fig_cifar}(a).

\textbf{Model-C-2}: The size of four base features are $32 \times 32, 16\times 16, 16\times 16, 8\times 8$. These four sub-networks corresponding to the base features have $8,6,4,2$ \textit{Conv Blocks}, respectively. Moreover, the numbers of input channels and the growth rates are $16, 32, 32, 64$ and $6, 12, 12, 24$, respectively. All \textit{Up-Conv Layers} are substituted to \textit{Regular Conv Layers} if the feature fusion happens between two same scales. The Model-C-2 has eight classifiers in total, and its overall architecture is illustrated in Figure \ref{fig_cifar}(b).

\textbf{Model-C-3}: The size of four base features are $32 \times 32, 16\times 16, 8\times 8, 8\times 8$. The numbers of input channels and the growth rates are $16, 16, 32, 64$ and $6, 6, 12, 24$, respectively. All \textit{Up-Conv Layers} are substituted to \textit{Regular Conv Layers} if the feature fusion happens between two same scales. The Model-C-3 has eight classifiers in total, and its overall architecture is illustrated in Figure \ref{fig_cifar}(c).

In the experiments, the Model-C-3 (even) are evaluated under the anytime classification setting (Figure 5 of the main paper), and all three models (lg) are evaluated under the budgeted batch classification setting (Figure 6 of the main paper).

\subsection{ImageNet}
\paragraph{MSDNet: } On the ImageNet, features with 4 different scales ($56\times 56, 28\times 28, 14\times 14, 7\times 7$) are used for MSDNets in our experiments. Three different MSDNets with five classifiers and  different depth are evaluated. Specifically, the $i^{th}$ classifier is attached at the $(t\times i + 3)^{th}$ layer where $i \in \{1, \cdots, 5\}$, and $t\in \{4, 6, 7\}$ is the step (number of layers) for each network block.

\paragraph{RANet: }The same 4 feature scales are utilized for our RANets in the experiments. The spatial resolutions of the base features are $56\times 56, 28\times 28, 14\times 14, 7\times 7$, respectively. We test 2 architecture configurations as follows:

\textbf{Model-I-1}: Four sub-networks corresponding to the base features have $8,6,4,2$ \textit{Conv Blocks}, respectively, and the number of layer in each \textit{Conv Block} is set to $8$. Moreover, the numbers of base feature channels and the growth rates are $32, 64, 64, 128$ and $16, 32, 32, 64$. For each \textit{Fusion Block}, compress factor of $0.25$ is applied. The Model-I-1 has eight classifiers in total, and its overall architecture is illustrated in Figure \ref{fig_ImageNet}.

\textbf{Model-I-2}: The architecture of the Model-I-2 is exactly the same as the Model-I-1. However, the numbers of base feature channels are $64, 128, 128, 256$.

In the experiments, the Model-I-2 is evaluated under the anytime classification setting (Figure 5 of the main paper), and both models are evaluated under the budgeted batch classification setting (Figure 6 of the main paper).

\section{Appendix B: Improved Techniques}
\begin{figure}[h]
    \begin{center}
    \includegraphics[width=0.45\textwidth]{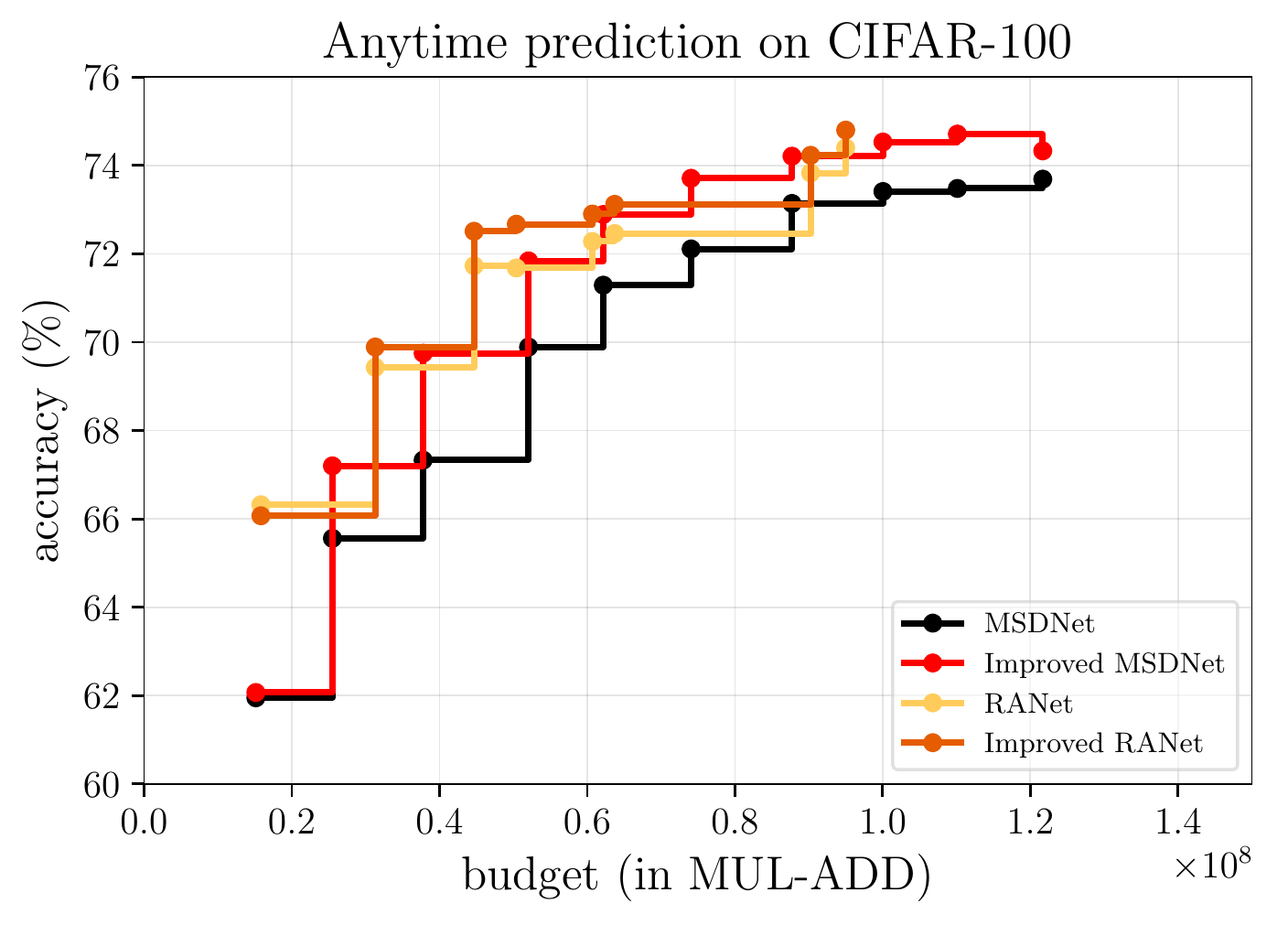}
    \end{center}
    \vskip -0.1in
    \caption{
        Accuracy (top-1) of anytime classification models as a function of average computational budget per image the on CIFAR-100, higher is better. MSDNet and RANet are trained with and without ISC and OFA techniques.
    }
    \label{fig_improved_anytime}
\end{figure}

\begin{figure}
    \begin{center}
    \includegraphics[width=0.45\textwidth]{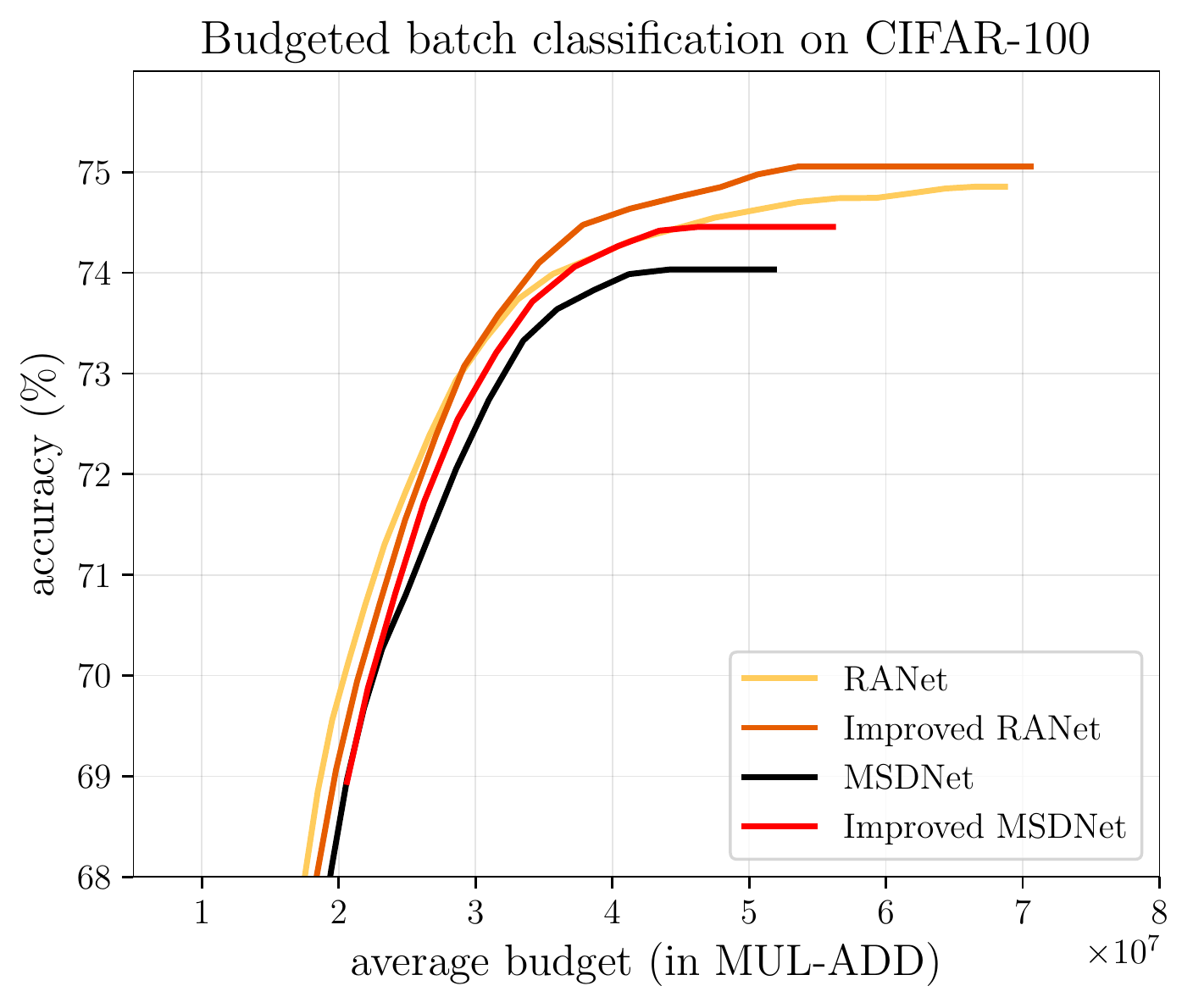}
    \end{center}
    \vskip -0.1in
    \caption{
        Accuracy (top-1) of budgeted batch classification models as a function of average computational budget per image the on CIFAR-100, higher is better. MSDNet and RANet are trained with and without ISC and OFA techniques.
    }
    \label{fig_improved}
\end{figure}

As some training techniques for adaptive inference models with multiple exits have been proposed in \cite{li2019improved}, we further evaluated the proposed RANet and MSDNet \cite{huang2017msdnet} with the implementation of these improved techniques on CIFAR-100. Inline Sub-network Collaboration (ISC) and One-For-All (OFA) knowledge distillation approaches are utilized in the experiments under anytime prediction and budgeted batch classification settings. Specifically, we implement these techniques (ISC and OFA) on our \textbf{Model-C-3} and MSDNet with 8 and 10 classifiers. The results are shown in Figure \ref{fig_improved_anytime} (anytime) and \ref{fig_improved} (budgeted batch).

For anytime prediction, Model-C-3 (even) and MSDNet with 10 classifiers are tested. From the results, we observe that the improved RANet can outperform the improved MSDNet, especially when the budget ranges from $0.3\times10^8$ to $0.6\times10^8$ FLOPs. Moreover, the improved RANet can achieve the highest accuracy ($~75\%$) with around $0.2\times10^8$ less FLOPs. We further observe that the techniques (ISC and OFA) do not work well on the first classifier of the RANet.

For budgeted batch classification, the results of RANet, Model-C-3 and MSDNet with 8 classifiers are tested. From the results, we observe that the improved RANet is still superior to the improved MSDNet, especially when the budget greater than $0.3\times10^8$. The original RANet can outperform the improved RANet can be due to the performance dropping of the first classifiers. However, compared with MSDNet and improved MSDNet, the accuracy of improved RANet can be $~1\%$ and  $~0.5\%$ higher respectively, which demonstrated the effectiveness of our RANet when implemented with the improved techniques.

{
   \small
   \bibliographystyle{ieee_fullname}
   \bibliography{RANetbib_abb}
}